\documentclass[11pt]{article}
\usepackage{acl2015}
\usepackage{times}
\usepackage{url}
\usepackage{latexsym}
\usepackage{arydshln}

\usepackage{graphicx}
\usepackage{amsmath, amsfonts, amssymb}

\title{Corpus-level Fine-grained Entity Typing Using Contextual Information}
\author{Yadollah Yaghoobzadeh \rm{and} \textbf{Hinrich Sch{\"u}tze}\\
  Center for Information and Language Processing \\
  University of Munich, Germany \\
  {\tt yadollah@cis.lmu.de} 
}

\date{ 14/08/2015}

\newcounter{notecounter}

\newcommand{\lowerpriorityenotesoff}{\long\gdef\lowerpriorityenote##1##2{}}
\newcommand{\lowerpriorityenoteson}{\long\gdef\lowerpriorityenote##1##2{{
\stepcounter{notecounter}
\large\bf
\hspace{1cm}\arabic{notecounter} $<<<$ ##1: ##2
$>>>$\hspace{1cm}}}}
\lowerpriorityenoteson
\lowerpriorityenotesoff

\newcommand{\enotesoff}{\long\gdef\enote##1##2{}}
\newcommand{\enoteson}{\long\gdef\enote##1##2{{
\stepcounter{notecounter}
\large\bf
\hspace{1cm}\arabic{notecounter} $<<<$ ##1: ##2
$>>>$\hspace{1cm}}}}
\enoteson
\enotesoff

\def\gmodel{global model}
\def\cmodel{context model}
\def\gmodelcap{Global model}
\def\cmodelcap{Context model}

\def\dnrm#1{\mbox{$_{\hbox{\scriptsize #1}}$}}

\def\tabref#1{Table~\ref{tab:#1}}

\def\secref#1{Section~\ref{sec:#1}}
\def\seclabel#1{\label{sec:#1}\label{p:#1}}
\def\eqref#1{Eq.~\ref{eqn:#1}}

\begin{document}
\maketitle
\begin{abstract}
This paper addresses the problem of corpus-level entity
typing, i.e., inferring from a large corpus that an entity
is a member of a class such as ``food'' or ``artist''.  
The application of entity typing we are interested in is
knowledge base completion, specifically, to learn which
classes an entity is a member of.
We
propose FIGMENT to tackle this
problem.  FIGMENT is embedding-based and combines (i) a \gmodel{} that
scores based on aggregated contextual information of an
entity and (ii) a \cmodel{} that first scores the individual
occurrences of an entity and then aggregates the scores.
In our evaluation, FIGMENT strongly outperforms 
an approach to entity typing 
that relies on relations obtained by an open
information extraction system.
\end{abstract}

\section{Introduction}
Natural language understanding (NLU) is not possible without
knowledge about the world -- partly so because world
knowledge is needed for many
NLP tasks that must be addressed as part of NLU; e.g.,
many coreference ambiguities can only be resolved based on
world knowledge. It is also true because most NLU
applications combine a variety of information sources that
include both text sources and knowledge bases; e.g.,
question answering systems need access to knowledge bases
like gazetteers. Thus, high-quality
knowledge bases are critical for successful NLU.

Unfortunately, most knowledge bases are
incomplete. The effort required to create knowledge bases is
considerable and since the world changes, it will always continue.
Knowledge bases are therefore always in need of updates and corrections. 
To address this problem, we
present an information extraction method that can be used
for knowledge base completion. 
In contrast to most other
work on knowledge base completion, we focus on fine-grained
\emph{classification of
entities} as opposed to \emph{relations between entities}.

\enote{hs}{i rewrote this paragraph for the following
  reasons

Methods in knowledge base completion can be categorized 
based on the information source they use.
Information extraction (IE) based methods use natural language corpora as a main source.
KB based methods use only internal KB information.
The aim of this paper is an IE-based method for entity classification.
To do so, our method must be \emph{corpus-level}, which means 
the inference over possible types of an entity must be based on 
all the mentions of the entity in the corpus.
This is different from the goal of \emph{sentence-level} 
methods, in which for each mention of an entity in the corpus
they must infer its locally possible types.

}

The goal of knowledge base completion is to acquire
knowledge in general as opposed to detailed analysis of an
individual context or sentence. Therefore, our approach is
\emph{corpus-level}: We infer the types of an entity 
by considering the set of all of
its mentions in the corpus.
In contrast, named entity recognition (NER) is 
\emph{context-level} or \emph{sentence-level}:
NER infers the type of an entity in a particular context. 
As will be discussed in more detail in the following
sections, the problems of  corpus-level entity typing vs.\
context/sentence-level entity typing are quite
different. This is partly because the objectives of
optimizing accuracy on the context-level vs.\ optimizing
accuracy on the corpus-level are different and partly
because evaluation measures for corpus-level and
context-level entity typing are different.

We define our problem as follows.
Let $K$ be a knowledge base that models a
set $E$ of entities, a set $T$ of fine-grained classes or \emph{types}
and a membership function $m:  E \times T \mapsto \{0,1\}$ such that
$m(e,t)=1$ iff entity $e$ has type $t$. Let $C$ be a large
corpus of text. Then, the problem we address in this paper
is \emph{corpus-level entity typing}:
For a given pair of entity $e$ and type $t$ determine -- based
on the evidence available in $C$ -- whether 
$e$ is a member of type $t$ (i.e., $m(e,t)=1$)
or not (i.e., $m(e,t)=0$)
and update the membership relation $m$
of $K$ with this information.

We investigate two approaches to entity typing:
a \gmodel{} and a \cmodel{}.

The \emph{\gmodel} aggregates all contextual information about an
entity $e$ from the corpus and then based on that, makes a classification
decision on a particular type $t$ -- i.e.,  
$m(e,t)=0$
vs.\ $m(e,t)=1$.

The \emph{\cmodel{}} first scores each individual context of $e$
as expressing type $t$ or not. A final decision on the value
of $m(e,t)$ is then made based on the distribution of
context scores.  One difficulty in knowledge base completion
based on text corpora is that it is too expensive to label
large amounts of text for supervised approaches.  For our
\cmodel{}, we address this problem using \emph{distant
  supervision}: we treat \emph{all} contexts of an entity that can
have type $t$ as contexts of type $t$ even though this
assumption will in general be only true for a \emph{subset} of these contexts.
Thus, as is typical for distant supervision, the labels
are incorrect in some contexts, but we will show that the
labeling is good enough to learn a high-quality \cmodel{}.

\enote{hs}{i changed this because the ``mention'' is not a
  ``context''

by considering entity mentions as contexts
of types

}

The \gmodel{} is potentially more robust since it looks
at all the available information at once. In contrast, the \cmodel{}
has the advantage that it can correctly predict types for
which there are only a small number of reliable
contexts. For example, in a large corpus we are likely to
find a few reliable contexts indicating that ``Barack Obama'' is
a bestselling author even though this evidence 
may be obscured in the global distribution
because the vast majority of mentions of ``Obama'' do not occur
in author contexts.


We implement the \gmodel{} and the \cmodel{} as well
as a simple combination of the two and call the resulting system
FIGMENT:
FIne-Grained eMbedding-based Entity Typing.
A key feature
of FIGMENT is that it makes extensive use of 
distributed vector representations
or \emph{embeddings}. 
We compute embeddings for words as is standard in a large
body of NLP literature, but we also compute embeddings for
\emph{entities} and for \emph{types}.
The motivation for using embeddings in these cases is
(i) better generalization 
and (ii) more robustness against noise for text types like web pages.
We compare the performance of
FIGMENT  with an approach based on Open Information Extraction (OpenIE).

The main contributions of this paper can be summarized as follows. 
\begin{itemize}
\item We address the problem of corpus-level entity typing
in a knowledge base completion setting.
In contrast to other
  work that has focused on learning \emph{relations} between
  entities, we learn \emph{types} of entities.
\item We show that context and \gmodel{}s for entity typing
  provide complementary information and combining them gives
  the best results.
\item We use embeddings for words, entities
  and types to improve generalization and deal with noisy input.
\item We show that our approach outperforms 
a  system based on OpenIE relations when the input corpus consists of
noisy web pages.
\end{itemize}


\enote{hs}{please rewrite the following: you are expected to
  refer to sections and their numbers}

In the following, we first discuss 
related work.
Then we motivate our approach and define the problem setting
we adopt.
We then introduce
our models in detail and report and analyze experimental results.
Finally,
we discuss remaining
challenges and possible future work and present our conclusions.

%
%
%
%
%
%
%
 
\section{Related work}\seclabel{related} 
\emph{Named entity recognition} (NER) 
is the task of
detecting and classifying named entities in text.  
While most NER systems (e.g., \newcite{finkel2005}) only
consider a small number of entity classes, recent work has addressed fine-grained NER
\cite{spaniol2012hyena,ling2012fine}. 
These methods use a variety of lexical and syntactic features to segment and classify
entity mentions.
Some more recent work assumes the segmentation is known and only classifies entity mentions. 
\newcite{donghybrid} use distributed representations  of words
in a hybrid classifier
to classify mentions to 20 types.
\newcite{yogatama2015acl} classify mentions 
to more fine-grained  types
by using different features for mentions and embedding labels in the same space. 
These methods as well as standard NER systems try to maximize correct classification of mentions in
individual contexts
whereas we aggregate individual contexts and evaluate on
accuracy of entity-type assignments inferred from the entire corpus.
In other words, their evaluation is \emph{sentence-level} 
whereas  ours is \emph{corpus-level}. 


\emph{Entity set expansion} (ESE)
is the problem of finding entities in a class (e.g.,
medications) given a seed set 
 (e.g., 
\{``Ibuprofen'',
``Maalox'', ``Prozac''\}).
The standard solution is pattern-based bootstrapping
\cite{thelen2002bootstrapping,gupta14evalpatterns}.  
ESE is different from the problem we address because ESE
starts with a small seed set whereas we assume that 
a large number of examples from a knowledge base (KB) is available. Initial experiments 
with the system of \newcite{gupta14evalpatterns} showed that
it was not performing well for our task -- this is not
surprising given that it is designed for a task with
properties quite different from entity typing.

More closely related to our work are 
the OpenIE systems NNPLB \cite{lin2012no} and PEARL
\cite{nakashole2013fine}
for 
\emph{fine-grained typing of unlinkable and emerging
entities}. Both systems first extract relation tuples from a
corpus and then type entities based on the tuples
they occur in (where NNPLB only uses the subject position
for typing). To perform typing, NNPLB propagates activation from known members
of a class to other entities whereas
PEARL assigns types to the argument slots of relations.
The main difference to FIGMENT is that we do not rely
on relation extraction. In principle, we can make use of any
context, not just subject and object positions. FIGMENT
also has advantages for noisy text for which relation extraction
can be challenging. This will be demonstrated in our
evaluation on web text. Finally, our emphasis is on making
yes-no decisions about possible types (as opposed to just ranking
possibilities) for all entities (as opposed to just emerging
or unlinkable entities). Our premise is that even existing
entities in KBs are often not completely modeled
and have entries that require enhancement.
We choose NNPLB as our baseline.


%


The fine-grained typing of entities performed by
FIGMENT can be used for \emph{knowledge base completion (KBC)}.
Most KBC systems focus on \emph{relations} between entities,
not on \emph{types} as we do.
Some  generalize the patterns of relationships within the KB \cite{nickel12yago,bordes2013transe} while  others
use a combination of within-KB generalization and
information extraction from text
\cite{Weston2013Conn,socher2013reasoning,Jiang12linkpred,Riedel13universal,Wang14joint}.
\newcite{neelakantan2015inferring} address entity typing in
a way that is
similar to FIGMENT.
Their method is based on KB information, more specifically entity descriptions in Wikipedia and Freebase. 
Thus, in contrast to our approach, their system is not able to type entities that are not
covered by existing KBs.
We infer
classes for entities from a large corpus and do not assume
that these entities occur in the KB.

%

\emph{Learning embeddings} for words is standard in a large
body of NLP literature (see  \newcite{baroni14predict} for
an overview). 
In addition to words, we also learn \emph{embeddings for entities and types}.
Most prior work on
entity embeddings 
(e.g., \newcite{Weston2013Conn}, \newcite{bordes2013transe})
and entity and type embeddings 
\cite{Zhao15relatedness}
has mainly used KB information as opposed
to text corpora.
\newcite{Wang14joint} learn embeddings
of words and entities in the same space by replacing
Wikipedia anchors with their corresponding entities.  For
our global model, we learn entity embedding in a similar
way, but on a corpus with automatically annotated entities.
For our context model, we learn and use type
embeddings jointly with corpus words to improve generalization, a novel contribution
of this paper to the best of our knowledge.
We learn all our embeddings using \texttt{word2vec}
 \cite{mikolov2013efficient}.

Our problem can be formulated as
\emph{multi-instance multi-label} (MIML) learning
\cite{zhou2006multi}, similar to the formulation for
relation extraction by \newcite{surdeanu2012multi}.  In our
problem, each example (entity) can have several instances
(contexts) and each instance can have several labels
(types).  Similar to
\newcite{zhou2006multi}'s work on  scene classification, we also
transform MIML into easier tasks.  The \gmodel{} transforms
MIML into a multi-label problem by merging all
instances of an example.  
The \cmodel{} solves the problem by combining the
instance-label scores to example-label scores.

\section{Motivation and problem definition}
\subsection{Freebase}
Large scale KBs like Freebase
\cite{bollacker2008freebase}, YAGO \cite{suchanek2007yago}
and Google knowledge graph are 
important NLP resources.  Their structure is
roughly equivalent to a graph in which entities are nodes
and edges are relations between entities. 
Each node is also associated with one or more semantic
classes, called types.
These types are the focus of this paper.


We use Freebase, the largest available KB, in this paper. 
In Freebase, an entity can
belong to several classes, e.g., ``Barack Obama'' is a member of 37 types including 
 ``US president'' and ``author''. 
One \emph{notable type} is also defined for each entity, e.g., 
``US-president'' for ``Obama'' since it is
regarded as his most prominent characteristic and the one that would
be used to disambiguate references to him, e.g., to distinguish him
from somebody else with the same name.

There are about
1500 types in Freebase, organized by domain; e.g.,
the domain ``food'' has types like ``food'', ``ingredient'' and
``restaurant''.
Some types 
like ``location'' are very general, some
are very fine-grained, e.g.,
``Vietnamese urban district''.
There are
types that have a large number of instances like
``citytown'' and types that have very few like ``camera\_sensor''.
Entities are
defined as instances of types.  They can have several types
based on the semantic classes that the entity they are
referring to is a member of --
as in the above example of Barack Obama.

The types are not organized in a strict taxonomy even though there exists an \textit{included type} relationship between types in Freebase.
The reason is that for a user-generated
KB it is difficult to maintain
taxonomic consistency. For example, almost all
instances of ``author'' are also instances of ``person'',
but sometimes organizations author and
publish documents. 
We follow the philosophy of Freebase and assume that the
types do not have a hierarchical organization.

\subsection{Incompleteness of knowledge base}
Even though Freebase is the largest publicly available
KB of its kind, it still has significant
coverage problems; e.g.,
78.5\% of persons in Freebase do not have
\textit{nationality} \cite{min2013distant}. 

\enote{hs}{deleted to save space

in our Freebase dump, 22\% of entities have only one type.  

}

This is unavoidable,
partly because Freebase is user-generated, partly because
the world changes and Freebase has to be updated to reflect
those changes.  All existing KBs that attempt to
model a large part of the world suffer from this
incompleteness problem.
Incompleteness is likely to become an even bigger problem in
the future as the number of types covered by KBs like
Freebase increases. As more and more fine-grained types are
added, achieving good coverage for these new
types using only human editors will become impossible.


The approach we adopt in this paper to address
incompleteness of KBs is extraction of information from
large text corpora. Text can be argued to be the main
repository of the type of knowledge represented in KBs, so
it is reasonable to attempt completing them based on
text. There is in fact a significant body of work on
corpus-based methods for extracting knowledge from text;
however, most of it has addressed relation extraction,
not the acquisition of type information -- roughly
corresponding to unary relations (see \secref{related}).  In
this paper, we focus on typing entities.

\subsection{Entity linking}
The first step in extracting information about entities from
text is to reliably identify mentions of these
entities.
This problem of \emph{entity linking} has some mutual
dependencies with entity typing.
Indeed,
some recent work shows large improvements when entity
typing and linking are jointly modeled
\cite{ling15entitylinking,durrett14entityanalysis}.
However, there are
constraints that are important for high-performance entity
linking, but that are of little relevance to entity typing.
For example, there is a large literature on entity linking
that deals with 
coreference resolution and
inter-entity constraints --
e.g., ``Naples'' is more likely to refer to a US (resp.\ an
Italian) city in a context mentioning ``Fort Myers''
(resp.\ ``Sicily'').  

Therefore, we will only address
entity typing in this paper
and
consider entity linking as an independent module 
that provides contexts of entities for FIGMENT.
More specifically, we  build FIGMENT on top of the output of an
existing entity linking system and use
FACC1,\footnote{\url{lemurproject.org/clueweb12/FACC1}} an automatic Freebase annotation of ClueWeb 
\cite{gabrilovich2013facc1}.
According to the FACC1 distributors,
precision of annotated entities is around 80-85\% and recall
is around 70-85\%.
\subsection{FIGER types}
Our goal is fine-grained typing of entities, but types like
``Vietnamese urban
district''
are too fine-grained.
To create a reliable setup for evaluation and to make sure
that all types have a reasonable number of instances, we
adopt the FIGER type set
\cite{ling2012fine} that was created with the same goals in mind.
FIGER consists of 112 tags and was created in an attempt to preserve the diversity of
Freebase types while consolidating infrequent and unusual
types through filtering and merging. For example,
the Freebase types 
``dish'', ``ingredient'', ``food''  and
``cheese''
are mapped to one type ``food''.  
See \cite{ling2012fine} for a complete list of  FIGER
types. We use ``type'' 
to refer to FIGER types in the rest of the paper.


\enote{hs}{i removed ``scoring'' from the section headline

or are you making a distinction between a ``model'' and a
``scoring model''?}

\section{Global, context and joint models}
\seclabel{models}
We address a problem setting in which the followings  are
given: a KB with a set of entities $E$, a set of types $T$
and 
a membership function 
$m:  E \times T \mapsto \{0,1\}$ such that
$m(e,t)=1$ iff entity $e$ has type $t$;
and a large annotated corpus $C$ in which mentions of 
$E$ are
linked. As mentioned before, we use FACC1 as our corpus.



In this problem setting, we address the task of
\emph{corpus-level fine-grained entity typing}: we want to
infer from the corpus for each pair of entity $e$ and 
type $t$ whether $m(e,t)=1$ holds, i.e., whether entity $e$ is a
member of type $t$. 

We use three scoring models in FIGMENT: \gmodel{},
\cmodel{} and joint model.  The models return a score
$S(e,t)$ for an entity-type pair $(e,t)$.  $S(e,t)$ is an
assessment of the extent to which it is true that the
semantic class $t$ contains $e$ and we learn it by training
on a subset of E.  The trained
models can  be applied to large corpora and the
resulting scores can be used for learning new types of
entities covered in the KB as well as for
typing new or unknown  entities  -- i.e., entities not
covered by the KB. To work for new or unknown entities, we would need
an entity linking system 
such as the ones participating in TAC KBP \cite{mcnamee2009overview}
that identifies and
clusters mentions of them.

\enote{yy}{
I changed these sentences, because I found it not clear enough to talk about
the typing of new entities rather than adding them to the KB.

covered by the KB. To add new entities to the KB, we would need
an entity tagging and linking system 
such as the ones participating in TAC KBP \cite{mcnamee2009overview}
that identifies and
clusters mentions of new entities.
}
\enote{hs}{you used ``connections'' above twice

i removed that because i find it confusing

sometimes it is necessary for a good writing to vary the way
you refer to a concept

i don't think it was necessary here

}

\subsection{\gmodelcap{}}
The  \gmodel{} (GM) scores possible types of entity
$e$ based on a \emph{distributed vector representation} or \emph{embedding}
$\vec{v}(e) \in \mathbb{R}^d$ of $e$.  
$\vec{v}(e)$ can be
learned from the entity-annotated corpus $C$.

Embeddings of words 
have been widely used in different
NLP applications.  
The embedding of a word is usually derived
from the distribution of its context words.
The hypothesis is that words with similar
meanings tend to occur in similar contexts \cite{harris54}
and therefore cooccur with similar context words.
By extension, the assumption of our model is that entities with similar types tend
to cooccur with similar context words.

\enote{hs}{still necessary? now discussed in realted work

footnote newcite{Wang14joint} also learns joint entity and word embeddings by mapping Wikipedia anchors to entities.

}


\enote{hs}{i didn't understand what you meant by ``proper''}

To learn a score function $S\dnrm{GM}(e,t)$, we use a
multilayer perceptron (MLP) with one shared hidden layer and
an output layer that contains, for each type $t$ in T, a
logistic regression classifier that predicts the
probability of $t$:
\begin{equation*} \label{eq:score_egm}
S\dnrm{GM}(e,t) = G_{t}\Big(\tanh\big(\textbf{W}\dnrm{input}\vec{v}(e)\big)\Big)
\end{equation*}
where 
$\textbf{W}\dnrm{input} \in \mathbb{R}^{h\times d} $ is the weight matrix from
$\vec{v}(e) \in \mathbb{R}^d$ to the hidden layer with size $h$. 
$G_{t}$ is the logistic regression classifier for type $t$ that is applied on the hidden layer.
  The shared hidden layer is designed to
exploit the dependencies among labels.
Stochastic gradient descent (SGD) with AdaGrad \cite{duchi2011adaptive} and minibatches are used to learn the parameters.

\enote{hs}{deleted: ``shared'' weight matrix}

%

\subsection{\cmodelcap{}}
\seclabel{ecm}
For the  \cmodel{} (CM), we first learn a scoring function
$S\dnrm{c2t}(c,t)$ for individual contexts $c$ in the
corpus.  $S\dnrm{c2t}(c,t)$ is an assessment of how likely
it is that an entity occurring in context $c$ has type
$t$. For example, 
consider the contexts $c_1$ =
``he served SLOT cooked in wine'' 
and $c_2$ = ``she loves SLOT more than anything''.
SLOT marks the
occurrence of an entity and it also shows that we do not
care about the entity mention itself but only its context.
For the type $t$ = ``food'',
$S\dnrm{c2t}(c_1,t)$ is high whereas 
$S\dnrm{c2t}(c_2,t)$ is low.
This example
demonstrates that some contexts of an entity like ``beef'' allow
specific inferences about its type whereas others do not.
We aim to learn a scoring function $S\dnrm{c2t}$ that can
distinguish these cases.


Based on the context scoring function $S\dnrm{c2t}$, we then
compute the corpus-level CM scoring function $S\dnrm{CM}$
that takes the scores $S\dnrm{c2t}(c_i,t)$ for all contexts
of entity $e$ in the corpus as input and returns a score
$S\dnrm{CM}(e,t)$ that assesses the appropriateness of $t$
for $e$. 
In other words, $S\dnrm{CM}$ is:
\begin{equation}
\label{eq:cmscore}
S\dnrm{CM}(e,t) = g(U_{e,t})
\end{equation}
where
$U_{e,t} = \{ S\dnrm{c2t}(c_1,t),$ $\ldots,
S\dnrm{c2t}(c_n,t)\}$ is the set of scores for $t$
based on the $n$ contexts $c_1 \ldots c_n$ of $e$ in the corpus.
The function $g$ is a summary function of the distribution
of scores, e.g., the mean, median or maximum. We use the mean in
this paper. 



We now describe how we learn $S\dnrm{c2t}$.  For training, we need
contexts that are labeled with types.
We do not have such a dataset in our problem setting, but we
can use the contexts of linked entities as distantly
supervised data.  Specifically, assume that entity
$e$ has $n$ types.  For each mention of $e$ in the corpus, we
generate a training example with $n$ labels, one for each of
the $n$
types of $e$.



\enote{hs}{is it not necessary to introduce the notion of
  unit here and then talk about units, not about words?}

For training $S\dnrm{c2t}$,
a context $c$ of a mention is represented
as the
concatenation of two vectors. 
One vector is the \emph{average} of the embeddings of the $2l$ words to the left and right of
the mention.
The other vector is the \emph{concatenation} of 
the embeddings of the $2k$ words to the left and right of the mention. 
E.g., for $k=2$ and $l=1$ the context $c$ is represented as the vector:  
$\Phi(c) = \big[x_{-2},x_{-1},x_{+1},x_{+2},\text{avg}(x_{-1},x_{+1})\big]$
where  $x_i \in \mathbb{R}^d$ is the embedding of the context word at
position $i$ relative to the entity in position $0$.

We train $S\dnrm{c2t}$ on
context representations that consist of embeddings
because our goal is a robust
model that works well on a wide variety of genres, including
noisy web pages. 
If there are other entities in the contexts, we first replace them with their notable type
to improve generalization. 
We learn word and type embeddings from the corpus C 
by replacing train entities with their notable type.

The next step is to score these examples. 
We use an MLP similar to the \gmodel{} to learn
$S\dnrm{c2t}$, which predicts the probability of type $t$
occurring in context $c$:
\begin{equation*}
\label{eq:c2teq}
S\dnrm{c2t}(c,t) = G_{t}\Big(\tanh\big(\textbf{W}\dnrm{input}\Phi(c)\big)\Big)
\end{equation*}
where
$\Phi(c) \in \mathbb{R}^{n}$ is the feature vector of the
context $c$ as described above,
$n=(2k+1)*d$ and
$\textbf{W}\dnrm{input} \in \mathbb{R}^{h\times n} $ is the weight matrix from input to hidden
layer with $h$ units.
Again, we use SGD with AdaGrad and minibatch training.

\enote{hs}{deleted: ``shared'' weight matrix}

\subsection{Joint model}
\gmodelcap{} and \cmodel{} have complementary strengths and weaknesses.

The strength of CM is that it is a direct model of the only
source of reliable evidence we have: the context in which
the entity occurs. This is also the way a human would
ordinarily do entity typing: she would determine if a
specific context in which the entity occurs implies that the
entity is, say, an author or a musician and type it
accordingly. The order of words is of critical importance
for the accurate assessment of a context and CM takes it
into account. A well-trained CM will also work for cases
for which GM is not applicable. In particular, if the KB
contains only a small number of entities of a particular
type, but the corpus contains a large number of contexts of
these entities, then CM is more likely to generalize
well.


The main weakness of CM is that a large proportion of
contexts does not contain sufficient information to infer
all types of the entity; e.g., based on our distant
supervised training data, we label every context of
``Obama'' with ``author'', ``politician'' and Obama's
other types in the KB.
Thus, CM is trained on a noisy training
set that contains only a relatively small number of
informative contexts.

The main strength of GM is that it 
bases its decisions on the entire evidence available in the
corpus. This makes it more robust. It is also more
efficient to train since its training set is by a factor of
$|M|$ smaller than the training set of CM where 
$|M|$ is the average number of  contexts per entity.

The disadvantage of GM is that it does not work well
for rare entities since the aggregated representation of an
entity may not be reliable if it is based on few
contexts.
It is also less likely to work well for non-dominant
types of an entity which might be swamped by dominant types;
e.g., the author contexts of ``Obama'' may be swamped by the
politician contexts and the overall context signature of the
entity ``Obama'' may not contain enough signal to infer that
he is an author.
Finally, methods for learning embeddings like
\texttt{word2vec} are bag-of-word  approaches.
Therefore, word order information -- 
critical for many typing decisions -- is lost.

Since GM and CM models are complementary, a combination
model should work better. We test this hypothesis for the
simplest possible joint model (JM), which adds the scores of the
two individual models:
\begin{equation*}
S\dnrm{JM}(e,t) = S\dnrm{GM}(e,t) + S\dnrm{CM}(e,t)
\end{equation*}

\section{Experimental setup and results}
\seclabel{exp}


\subsection{Setup}
\textbf{Baseline:} 
Our baseline system is the OpenIE system no-noun-phrase-left-behind (NNPLB)  by
\newcite{lin2012no} (see
\secref{related}).
Our reimplementation performs on a par with published results.\footnote{The precision of our implementation
  on the dataset of three million relation triples
  distributed by \cite{lin2012no}
is 60.7\% compared to
  59.8\% and 61\% for tail and head entities
  reported by \newcite{lin2012no}.}
We use NNPBL as an alternative way of computing scores
$S(e,t)$. 
Scores of the four systems we compare --
NNPBL, GM, CM, JM -- are processed the same way to
perform entity typing (see below).

\enote{hs}{without the last sentence, is it clear how the
  baseline is used? or is the sentence incorrect?
}

\textbf{Corpus:}
We select a subset of about 7.5
million web pages, taken from the first segment of
ClueWeb12,\footnote{\url{http://lemurproject.org/clueweb12}} from different crawl types: 1 million Twitter links, 120,000 WikiTravel pages and 6.5 million web pages.
This corpus is preprocessed by eliminating HTML tags,
replacing all numbers with ``7'' and all web links and email
addresses with ``HTTP'', filtering out sentences with length
less than 40 characters, and finally doing a simple tokenization.  
We  merge the text with the
FACC1 annotations.
The resulting
corpus has  4 billion tokens and  950,000
distinct entities. We use
the 2014-03-09 
Freebase data dump 
as our  KB.

\enote{hs}{is htis important?

300 million lines,

}

\textbf{Entity datasets:}
We consider all entities in the corpus  whose notable types can be mapped 
to one of the 112 FIGER types, based on the mapping provided by FIGER. 
750,000 such entities form our set of entities.
10 out of 112 FIGER types have no entities in this
set.\footnote{The reason is that
the FIGER mapping uses Freebase user-created
classes. The 10 missing types are not
the notable type of any entity in Freebase.}

We run the OpenIE system Reverb \cite{ReVerb2011} to extract
relation triples
of
the form $<$subject, relation, object$>$. 
Since NNPLB  only considers entities 
in the subject position,
we filter out triples whose subject is not an entity.
The size of the remaining set of triples is 4,000,000.
For a direct
comparison with NNPLB, we divide the 750,000 entities into those that
occur in subject position in one of the extracted triples
(about 250,000 \emph{subject entities} or SE) and those that
do not (about 500,000 \emph{non-subject entities} or NSE).
We split SE 50:20:30 into train, dev and test sets.  
The average and median number of FIGER types of the training
entities  are 1.8 and 2, respectively. 
We use NSE to evaluate performance of FIGMENT on entities
that do not occur in subject position.\footnote{The entity datasets are available at
\url{http://cistern.cis.lmu.de/figment}
}

\textbf{Context sampling:} 
For $S\dnrm{c2t}$,
we create train', dev' and
test' sets of \emph{contexts} that correspond to 
train, dev and test sets of \emph{entities}. 
Because the number of contexts is unbalanced for both
entities and types and because we want to
accelerate training and testing, we downsample contexts.
For the set train', we use the notable type feature of Freebase:
For each type $t$, we take contexts from the mentions of
those entities whose notable type is $t$.
Recall, however, that each context is labeled with all types of its
entity -- see
\secref{ecm}.

Then if the number of contexts for $t$ is larger than a minimum, 
we sample the contexts based on the number of training entities of $t$. 
We set the minimum to 10,000 and constrain the number of samples for each $t$ to 20,000. 
Also, to reduce the effect of distant supervision, entities with fewer distinct types
are preferred in sampling to provide discriminative contexts for their notable types.
For test' and dev' sets, we sample 300 and 200 random contexts, respectively, for each entity.

\textbf{System setup:}
As the baseline, we apply NNPLB to the 4 million extracted triples.
To learn entity embeddings for GM,
we run \texttt{word2vec}
(skipgram, 200
dimensions, window size 5) on a version of the corpus in which
entities have been replaced by their Freebase IDs,
based on the FACC1 annotation. 
We then train MLP with number of hidden units $h$ = 200 
on the embeddings of training entities
until the error on dev entities stops
decreasing. 

Our reasoning for the unsupervised training setup is that we
do not use any information about the types of entities
(e.g., no entities annotated by humans with types) when we run an unsupervised
algorithm like \texttt{word2vec}. 
In a real-world application of
FIGMENT to a new corpus, we would first run \texttt{word2vec} on the
merger of our corpus and the new
corpus, retrain GM  on training entities and finally apply it to
entities in the new corpus. This scenario is simulated by
our setup.

\enote{hs}{isn't this description inconsitent with the prior description?}

Recall that the input to CM consists of $2k$ unit embeddings and
the average of
$2l$ unit embeddings where we use the term
\emph{unit} to refer to both words and types.  We set $k$ to
4 and $l$ to 5.  To learn embeddings for units, we first
exclude lines containing test entities, and then replace
each entity with its notable type.  Then, we run
\texttt{word2vec} (skipgram, 100 dimensions, window size 5)
on this new corpus and learn embeddings for words and
types.

Using the embeddings as input representations,  
we train $S\dnrm{c2t}$ on train' until error on dev' stops
decreasing. We set the number of hidden units to 300.
We then apply the trained scoring function
$S\dnrm{c2t}$ to
test'  and get the scores $S\dnrm{c2t}(c,t)$ for test' contexts.
As explained in  \secref{ecm},
we compute the corpus-level scores $S\dnrm{CM}(e,t)$ for each entity by averaging
its context-level scores (see Equation \ref{eq:cmscore}).


\textbf{Ranking evaluation:}
This evaluation shows how well the models rank types for entities.
The ranking is based on the scores
$S(e,t)$ produced by the different models and baselines. 
Similar to the evaluation performed by \newcite{lin2012no},
we use precision at 1 (P@1) and breakeven point (BEP, \newcite{boldrin2008against}).
BEP is $F_1$ at the point in the ranked list at
which precision and recall have the same value.

\textbf{Classification evaluation:}
\seclabel{class}
This evaluation demonstrates the quality of the thresholded assignment decisions
produced by the models. 
These measures more directly express how well
FIGMENT would succeed in enhancing the KB with new
information since for each pair $(e,t)$, we have to make a
binary decision about whether to put it in the KB or not.
We compare our decisions with the gold KB information. 

Our evaluation measures are 
(i) accuracy: an entity is correct if
all its types and no incorrect types are assigned to it;
(ii) micro average: $F_1$ of all type-entity assignment decisions;
(iii) entity macro average $F_1$: $F_1$ of types assigned to
an entity, averaged over entities; 
(iv) type macro average $F_1$: $F_1$ of entities assigned to
a type, averaged over types.

The assignment decision is made based on thresholds, one
 per type, for each $S(e,t)$. 
We select
the threshold that maximizes 
$F_1$ of entities assigned
to the type on dev.

%
%
%

\begin{table*}[tbs]
\begin{center}
\setlength{\tabcolsep}{3pt}
{
\begin{tabular}{l|cc|ccc|cc|ccc|cc|ccc}
& \multicolumn{5}{|c|}{all entities} &
  \multicolumn{5}{|c|}{head entities} &
  \multicolumn{5}{|c}{tail entities}\\
& P@1 & BEP & acc & mic &  mac 
& P@1 & BEP & acc &  mic &  mac 
& P@1 & BEP & acc &  mic &  mac\\ 
\hline %
MFT    & .101 & .406 &  -   & -    &  -  
	   & .111 & .410 &  -   &  -   &  -   
	   & .097 & .394 &  -   &  -   &  -  \\
NNPLB  & .365 & .480 & .000 & .099 & .096
       & .378 & .503 & .000 & .114 & .109 
       & .368 & .474 & .000 & .086 & .084\\
CM     & .694 & .734 & .299 & .668 & .635 
	   & .713 & .751 & .385 & .738 & .702 
	   & .608 & .661 & .118 & .487 & .452\\  
GM     & .805 & .856 & .426 & .733 & .688 
       & .869 & .899 & .489 & .796 & .769 
       & .665 & .757 & .299 & .578 & .510\\ 
JM     & \textbf{.816} & \textbf{.860} & \textbf{.435} & \textbf{.743} & \textbf{.699} 
       & \textbf{.874  }& \textbf{.900} & \textbf{.500 }& \textbf{.803} & \textbf{.776} 
       & \textbf{.688} & \textbf{.764} & \textbf{.306} & \textbf{.601} & \textbf{.532}
\end{tabular} 
}
\caption{Ranking and classification results for SE entities. 
P@1 and BEP are ranking measures. 
Accuracy (acc),  micro (mic) and macro (mac) are  classification measures.}\label{tab:bigtable}
\end{center}
\end{table*}


\begin{table}[tbp]
\begin{center}
{
\begin{tabular}{l|c|c|c}
       & all types   & head types & tail types\\ 
\hline
NNPLB  & .092 & .246       & .066\\
CM     & .406 & .662       & .268\\ 
GM     & .533 & .725       & .387\\ 
JM     &\textbf{.545} & \textbf{.734}       & \textbf{.407} 
\end{tabular} 
}
\caption{Type macro average $F_1$ for all, head and tail types}
\label{tab:table3FIG}
\end{center}
\end{table}

\subsection{Results}
\tabref{bigtable} presents results for
the ranking evaluation as
well as for the first three measures of the classification
evaluation.  MFT is the most frequent type baseline that
ranks types according to their frequency in train.  We also
show the results for head entities (frequency higher than
100) and tail entities (frequency less than 5).  The
performance of the systems is in this order: JM $>$ GM $>$
CM $>$ NNPLB $>$ MFT.


\tabref{table3FIG} shows the results of the fourth classification measure, type macro average $F_1$, for all,
head (more than 3000 train entities, 11 types), and tail
(less than 200 train entities, 36 types) types.  The
ordering of models for \tabref{table3FIG} is in
line with \tabref{bigtable}:  JM $>$  GM
$>$ CM $>$ NNPLB $>$ MFT.


\enote{hs}{orphans should be avoided}

We can easily run FIGMENT for \textbf{non-subject entities} (NSE)
exactly the same way we have run it for subject entities.
We test our JM on 
the 67,000 NSE entities with a
frequency of more than 10. 
The top ranked type returned for 73.5\% of entities was correct. 
Thus, due to our ability to
deal with NSE, we can type an additional 50,000 entities correctly.


\begin{table}[tbp]
\setlength{\tabcolsep}{3pt}
{

\begin{center}
\begin{tabular}{c|c@{\hspace{0.05cm}}c@{\hspace{0.05cm}}c@{\hspace{0.05cm}}c@{\hspace{0.05cm}}c@{\hspace{0.05cm}}c@{\hspace{0.05cm}}c@{\hspace{0.05cm}}c}
2k    &  0   &  2   &   4  &  6   & 8    &  10    &   12     &     14\\ 
h     &  50  & 100  &  200 & 250  & 300   & 400   &   450    &     450\\
\hline 
micro & .576 & .613 & .672 & .673 & .668 & .674 & \textbf{.680} & .674    \\ 
P@1   & .663 & .685 & .687 & .718 & .694 & \textbf{.744} & .722 & .742
\end{tabular} 
\caption{Effect of the context size $2k$ in CM (2k: context size, h: number of hidden units in MLP)}
\label{tab:contextsize}
\end{center}
}

\end{table}

\section{Analysis}
\seclabel{analysis}
\textbf{Effect of window size in CM:}
\tabref{contextsize}
explores the effect of using different context sizes.
Recall that CM was trained with $2k$ = 8 for the concatenation
and $2l$ = 10 for the average window size. 
We change $2k$ from 0 to 14 while keeping $2l$ = 10.
The number of hidden units used in each model is also reported. 
The table shows that CM can leverage larger context sizes well.

\textbf{Poor results of NNPLB:} 
NNPLB is mostly hampered by Reverb, which did not work well on the noisy web corpus. 
As a result, the quality of the extracted relations -- which
NNPLB entity typing is based on -- is too low 
for reliable typing decisions. 
The good results of NNPLB on their non-noisy published
relation triples confirm that. On the three million relation triples,  
when mapping Freebase types to FIGER, 
 P@1 of NNPLB is .684; when limiting  entities to those with more
than 10 relations, the results improve to .776.

\textbf{GM performs better than CM and JM performs best:} 
The fact that GM outperforms CM shows that decisions based on one 
global vector of an entity work better than aggregating multiple weak decisions on their
contexts. That is clearest for tail entities -- where  one
bad context can highly influence the final decision -- and for
tail types, which CM  was not able to 
distinguish from other similar types.  However,
the good results of the simple JM confirm that
the score distributions in CM do help.  As an example,
consider one of the 
test entities that is an ``author''. GM and CM wrongly
predict ``written\_work'' and ``artist'', respectively, but
JM correctly outputs ``author''.

%

\textbf{Errors of CM:} 
Many CM errors are caused by its simple input representation: 
it has to learn all linguistic abstractions that it wants to
rely on from the training set. One manifestation of this
problem is that CM confuses the types ``food'' and
``restaurant''. There are only few linguistic contexts in which
entities of these types can be exchanged for each other.
 On the other hand, the context words
they cooccur with in a bag-of-words (BOW) sense are very
similar. Thus, this indicates that CM pays too much
attention to BOW information and that its 
representation of contexts is limited in terms of
generalization.

\textbf{Assumptions that result in errors:} 
The performance of all models suffers from a number of
assumptions we made in our training / evaluation setup that 
are only approximately true.

The first assumption is that FACC1 is correct. But it has
a precision of only 80-85\% and this caused many errors.
An example is the lunar crater
``Buffon'' in Freebase, a ``location''. 
Its predicted type is ``athlete'' because 
some FACC1 annotations of the crater link it to the Italian goalkeeper.

The second assumption of our evaluation setup is the completeness of Freebase.
There are about 2,600 entities with the single
type ``person'' in SE test.
For 62\% of the errors on this
subset, the top predicted type is a subtype of person:
``author'', ``artist'' etc.  We manually typed a random
subset of 50 and found that the
predicted type is actually correct for 44 of these entities.

The last assumption is the mapping from Freebase to FIGER.
Some common Freebase types like ``award-winner'' are not
mapped.  This negatively affects evaluation measures for
many entities. On the other hand, the resulting types do not
have a balanced number of instances. Based on our training
entities, 11 types (e.g., ``law'') have
less than 50 instances while 26 types (e.g., 
``software'') have more than 1000 instances.  Even
sampling the contexts could not resolve this problem and this led to
low performance on tail types.


\section{Future work}
The performance of FIGMENT is poor for tail types and
entities. We plan to address this in the future  (i)
by running FIGMENT on larger corpora, (ii)
by refining
the FIGER type set to cover more Freebase entities, (iii)
by  exploiting a hierarchy over types and (iv)
by exploring more complex input representations of the
context for the CM.


FIGMENT's context model can in principle be based on any
system that provides entity-type assessment scores for
individual contexts. Thus, 
as an alternative to our scoring model $S\dnrm{c2t}(c,t)$,
we could use
sentence-level entity classification systems such as
FIGER \cite{ling2012fine} and
\cite{yogatama2015acl}'s system.
These systems are based on  linguistic features different
from the input representation we use, so
a comparison with our embedding-based approach is interesting.
Our assumption is that FIGMENT is more robust against noise,
but investigation is needed. 

The components of 
the version of FIGMENT we presented, in particular,
 \cmodel{} and
\gmodel{},
do not use  features 
derived from the mention of an entity.
Our assumption was that such features 
are less useful
for fine-grained entity typing. However, there are clearly
some types for which mention-based features are useful
(e.g., medications or organizations referred to by
abbreviations), so we will investigate the usefulness of
such features in the future.

\enote{hs}{i removed ``scalable'': are you sure thi swould
  be a scalability issue

Our assumption was that a scalable model, which is able to do more and more fine-grained typing, 
should not rely on individual entity
features (like name patterns).
However, an experiment is necessary to justify our assumption.

}


\section{Conclusion}
We presented FIGMENT, 
a corpus-level system
that uses contextual information for entity typing.
We designed two scoring models for pairs of entities and types: 
a \gmodel{} that scores based
on aggregated context information 
and a \cmodel{} that aggregates the scores of individual contexts.
We used embeddings of words, entities and types to 
represent  contextual information.
Our experimental results show that \gmodel{} and  \cmodel{} provide
complementary information for entity typing.
We demonstrated that, 
comparing with an OpenIE-based system, 
FIGMENT performs well on noisy web pages.

\enote{hs}{can this be cut to save space?

We experimented FIGMENT 
to rank types for entities and to do entity type classification. 

}

\paragraph{Acknowledgements.}
Thanks to the anonymous reviewers for their valuable comments.
This work was supported by 
Deutsche Forschungsgemeinschaft (grant DFG SCHU 2246/8-2,
SPP 1335).

\newpage

\bibliographystyle{acl}
\bibliography{emnlp2015}

\begin{thebibliography}{}

\bibitem[\protect\citename{Baroni \bgroup et al.\egroup }2014]{baroni14predict}
Marco Baroni, Georgiana Dinu, and Germ{\'{a}}n Kruszewski.
\newblock 2014.
\newblock Don't count, predict! {A} systematic comparison of context-counting
  vs. context-predicting semantic vectors.
\newblock In {\em Proceedings of the 52nd Annual Meeting of the Association for
  Computational Linguistics, {ACL} 2014}, pages 238--247.

\bibitem[\protect\citename{Boldrin and Levine}2008]{boldrin2008against}
Michele Boldrin and David~K. Levine.
\newblock 2008.
\newblock {\em Against intellectual monopoly}.
\newblock Cambridge University Press Cambridge.

\bibitem[\protect\citename{Bollacker \bgroup et al.\egroup
  }2008]{bollacker2008freebase}
Kurt~D. Bollacker, Colin Evans, Praveen Paritosh, Tim Sturge, and Jamie Taylor.
\newblock 2008.
\newblock Freebase: a collaboratively created graph database for structuring
  human knowledge.
\newblock In {\em Proceedings of the {ACM} {SIGMOD} International Conference on
  Management of Data, {SIGMOD} 2008, Vancouver, BC, Canada, June 10-12, 2008},
  pages 1247--1250.

\bibitem[\protect\citename{Bordes \bgroup et al.\egroup
  }2013]{bordes2013transe}
Antoine Bordes, Nicolas Usunier, Alberto Garc{\'{\i}}a{-}Dur{\'{a}}n, Jason
  Weston, and Oksana Yakhnenko.
\newblock 2013.
\newblock Irreflexive and hierarchical relations as translations.
\newblock {\em CoRR}, abs/1304.7158.

\bibitem[\protect\citename{Dong \bgroup et al.\egroup }2015]{donghybrid}
Li~Dong, Furu Wei, Hong Sun, Ming Zhou, and Ke~Xu.
\newblock 2015.
\newblock A hybrid neural model for type classification of entity mentions.
\newblock In {\em Proceedings of the Twenty-Fourth International Joint
  Conference on Artificial Intelligence, {IJCAI} 2015, Buenos Aires, Argentina,
  July 25-31, 2015}, pages 1243--1249.

\bibitem[\protect\citename{Duchi \bgroup et al.\egroup
  }2011]{duchi2011adaptive}
John~C. Duchi, Elad Hazan, and Yoram Singer.
\newblock 2011.
\newblock Adaptive subgradient methods for online learning and stochastic
  optimization.
\newblock {\em Journal of Machine Learning Research}, 12:2121--2159.

\bibitem[\protect\citename{Durrett and Klein}2014]{durrett14entityanalysis}
Greg Durrett and Dan Klein.
\newblock 2014.
\newblock A joint model for entity analysis: Coreference, typing, and linking.
\newblock {\em {TACL}}, 2:477--490.

\bibitem[\protect\citename{Fader \bgroup et al.\egroup }2011]{ReVerb2011}
Anthony Fader, Stephen Soderland, and Oren Etzioni.
\newblock 2011.
\newblock Identifying relations for open information extraction.
\newblock In {\em Proceedings of the 2011 Conference on Empirical Methods in
  Natural Language Processing, {EMNLP} 2011, 27-31 July 2011, John McIntyre
  Conference Centre, Edinburgh, UK, {A} meeting of SIGDAT, a Special Interest
  Group of the {ACL}}, pages 1535--1545.

\bibitem[\protect\citename{Finkel \bgroup et al.\egroup }2005]{finkel2005}
Jenny~Rose Finkel, Trond Grenager, and Christopher~D. Manning.
\newblock 2005.
\newblock Incorporating non-local information into information extraction
  systems by gibbs sampling.
\newblock In {\em {ACL} 2005, 43rd Annual Meeting of the Association for
  Computational Linguistics, Proceedings of the Conference, 25-30 June 2005,
  University of Michigan, {USA}}.

\bibitem[\protect\citename{Gabrilovich \bgroup et al.\egroup
  }2013]{gabrilovich2013facc1}
Evgeniy Gabrilovich, Michael Ringgaard, and Amarnag Subramanya.
\newblock 2013.
\newblock Facc1: Freebase annotation of clueweb corpora.

\bibitem[\protect\citename{Gupta and Manning}2014]{gupta14evalpatterns}
Sonal Gupta and Christopher~D. Manning.
\newblock 2014.
\newblock Improved pattern learning for bootstrapped entity extraction.
\newblock In {\em Proceedings of the Eighteenth Conference on Computational
  Natural Language Learning, CoNLL 2014, Baltimore, Maryland, USA, June 26-27,
  2014}, pages 98--108.

\bibitem[\protect\citename{Harris}1954]{harris54}
Zellig~S. Harris.
\newblock 1954.
\newblock Distributional structure.
\newblock {\em Word}, 10:146--162.

\bibitem[\protect\citename{Jiang \bgroup et al.\egroup }2012]{Jiang12linkpred}
Xueyan Jiang, Volker Tresp, Yi~Huang, and Maximilian Nickel.
\newblock 2012.
\newblock Link prediction in multi-relational graphs using additive models.
\newblock In {\em Proceedings of the International Workshop on Semantic
  Technologies meet Recommender Systems {\&} Big Data, Boston, USA, November
  11, 2012}, pages 1--12.

\bibitem[\protect\citename{Lin \bgroup et al.\egroup }2012]{lin2012no}
Thomas Lin, Mausam, and Oren Etzioni.
\newblock 2012.
\newblock No noun phrase left behind: Detecting and typing unlinkable entities.
\newblock In {\em Proceedings of the 2012 Joint Conference on Empirical Methods
  in Natural Language Processing and Computational Natural Language Learning,
  EMNLP-CoNLL 2012, July 12-14, 2012, Jeju Island, Korea}, pages 893--903.

\bibitem[\protect\citename{Ling and Weld}2012]{ling2012fine}
Xiao Ling and Daniel~S. Weld.
\newblock 2012.
\newblock Fine-grained entity recognition.
\newblock In {\em Proceedings of the Twenty-Sixth {AAAI} Conference on
  Artificial Intelligence, July 22-26, 2012, Toronto, Ontario, Canada.}

\bibitem[\protect\citename{Ling \bgroup et al.\egroup
  }2015]{ling15entitylinking}
Xiao Ling, Sameer Singh, and Daniel~S. Weld.
\newblock 2015.
\newblock Design challenges for entity linking.
\newblock {\em {TACL}}, 3:315--328.

\bibitem[\protect\citename{McNamee and Dang}2009]{mcnamee2009overview}
Paul McNamee and Hoa~Trang Dang.
\newblock 2009.
\newblock Overview of the tac 2009 knowledge base population track.
\newblock In {\em Text Analysis Conference (TAC)}, volume~17, pages 111--113.

\bibitem[\protect\citename{Mikolov \bgroup et al.\egroup
  }2013]{mikolov2013efficient}
Tomas Mikolov, Kai Chen, Greg Corrado, and Jeffrey Dean.
\newblock 2013.
\newblock Efficient estimation of word representations in vector space.
\newblock {\em CoRR}, abs/1301.3781.

\bibitem[\protect\citename{Min \bgroup et al.\egroup }2013]{min2013distant}
Bonan Min, Ralph Grishman, Li~Wan, Chang Wang, and David Gondek.
\newblock 2013.
\newblock Distant supervision for relation extraction with an incomplete
  knowledge base.
\newblock In {\em Human Language Technologies: Conference of the North American
  Chapter of the Association of Computational Linguistics, Proceedings, June
  9-14, 2013, Westin Peachtree Plaza Hotel, Atlanta, Georgia, {USA}}, pages
  777--782.

\bibitem[\protect\citename{Nakashole \bgroup et al.\egroup
  }2013]{nakashole2013fine}
Ndapandula Nakashole, Tomasz Tylenda, and Gerhard Weikum.
\newblock 2013.
\newblock Fine-grained semantic typing of emerging entities.
\newblock In {\em Proceedings of the 51st Annual Meeting of the Association for
  Computational Linguistics, {ACL} 2013, 4-9 August 2013, Sofia, Bulgaria,
  Volume 1: Long Papers}, pages 1488--1497.

\bibitem[\protect\citename{Neelakantan and
  Chang}2015]{neelakantan2015inferring}
Arvind Neelakantan and Ming{-}Wei Chang.
\newblock 2015.
\newblock Inferring missing entity type instances for knowledge base
  completion: New dataset and methods.
\newblock In {\em {NAACL} {HLT} 2015, The 2015 Conference of the North American
  Chapter of the Association for Computational Linguistics: Human Language
  Technologies, Denver, Colorado, USA, May 31 - June 5, 2015}, pages 515--525.

\bibitem[\protect\citename{Nickel \bgroup et al.\egroup }2012]{nickel12yago}
Maximilian Nickel, Volker Tresp, and Hans{-}Peter Kriegel.
\newblock 2012.
\newblock Factorizing {YAGO:} scalable machine learning for linked data.
\newblock In {\em World Wide Web Conference}, pages 271--280.

\bibitem[\protect\citename{Riedel \bgroup et al.\egroup
  }2013]{Riedel13universal}
Sebastian Riedel, Limin Yao, Andrew McCallum, and Benjamin~M. Marlin.
\newblock 2013.
\newblock Relation extraction with matrix factorization and universal schemas.
\newblock In {\em Human Language Technologies: Conference of the North American
  Chapter of the Association of Computational Linguistics, Proceedings, June
  9-14, 2013, Westin Peachtree Plaza Hotel, Atlanta, Georgia, {USA}}, pages
  74--84.

\bibitem[\protect\citename{Socher \bgroup et al.\egroup
  }2013]{socher2013reasoning}
Richard Socher, Danqi Chen, Christopher~D. Manning, and Andrew~Y. Ng.
\newblock 2013.
\newblock Reasoning with neural tensor networks for knowledge base completion.
\newblock In {\em Advances in Neural Information Processing Systems 26: 27th
  Annual Conference on Neural Information Processing Systems 2013. Proceedings
  of a meeting held December 5-8, 2013, Lake Tahoe, Nevada, United States.},
  pages 926--934.

\bibitem[\protect\citename{Suchanek \bgroup et al.\egroup
  }2007]{suchanek2007yago}
Fabian~M. Suchanek, Gjergji Kasneci, and Gerhard Weikum.
\newblock 2007.
\newblock Yago: a core of semantic knowledge.
\newblock In {\em Proceedings of the 16th International Conference on World
  Wide Web, {WWW} 2007, Banff, Alberta, Canada, May 8-12, 2007}, pages
  697--706.

\bibitem[\protect\citename{Surdeanu \bgroup et al.\egroup
  }2012]{surdeanu2012multi}
Mihai Surdeanu, Julie Tibshirani, Ramesh Nallapati, and Christopher~D. Manning.
\newblock 2012.
\newblock Multi-instance multi-label learning for relation extraction.
\newblock In {\em Proceedings of the 2012 Joint Conference on Empirical Methods
  in Natural Language Processing and Computational Natural Language Learning,
  EMNLP-CoNLL 2012, July 12-14, 2012, Jeju Island, Korea}, pages 455--465.

\bibitem[\protect\citename{Thelen and Riloff}2002]{thelen2002bootstrapping}
Michael Thelen and Ellen Riloff.
\newblock 2002.
\newblock A bootstrapping method for learning semantic lexicons using
  extraction pattern contexts.
\newblock In {\em Proceedings of the ACL-02 Conference on Empirical Methods in
  Natural Language Processing, EMNLP 2002, Stroudsburg, PA, USA}, pages
  214--221.

\bibitem[\protect\citename{Wang \bgroup et al.\egroup }2014]{Wang14joint}
Zhen Wang, Jianwen Zhang, Jianlin Feng, and Zheng Chen.
\newblock 2014.
\newblock Knowledge graph and text jointly embedding.
\newblock In {\em Proceedings of the 2014 Conference on Empirical Methods in
  Natural Language Processing, {EMNLP} 2014, October 25-29, 2014, Doha, Qatar,
  {A} meeting of SIGDAT, a Special Interest Group of the {ACL}}, pages
  1591--1601.

\bibitem[\protect\citename{Weston \bgroup et al.\egroup }2013]{Weston2013Conn}
Jason Weston, Antoine Bordes, Oksana Yakhnenko, and Nicolas Usunier.
\newblock 2013.
\newblock Connecting language and knowledge bases with embedding models for
  relation extraction.
\newblock In {\em Proceedings of the 2013 Conference on Empirical Methods in
  Natural Language Processing, {EMNLP} 2013, 18-21 October 2013, Grand Hyatt
  Seattle, Seattle, Washington, USA, {A} meeting of SIGDAT, a Special Interest
  Group of the {ACL}}, pages 1366--1371.

\bibitem[\protect\citename{Yogatama \bgroup et al.\egroup
  }2015]{yogatama2015acl}
Dani Yogatama, Daniel Gillick, and Nevena Lazic.
\newblock 2015.
\newblock Embedding methods for fine grained entity type classification.
\newblock In {\em Proceedings of the 53rd Annual Meeting of the Association for
  Computational Linguistics and the 7th International Joint Conference on
  Natural Language Processing of the Asian Federation of Natural Language
  Processing, {ACL} 2015, July 26-31, 2015, Beijing, China, Volume 2: Short
  Papers}, pages 291--296.

\bibitem[\protect\citename{Yosef \bgroup et al.\egroup }2012]{spaniol2012hyena}
Mohamed~Amir Yosef, Sandro Bauer, Johannes Hoffart, Marc Spaniol, and Gerhard
  Weikum.
\newblock 2012.
\newblock {HYENA:} hierarchical type classification for entity names.
\newblock In {\em {COLING} 2012, 24th International Conference on Computational
  Linguistics, Proceedings of the Conference: Posters, 8-15 December 2012,
  Mumbai, India}, pages 1361--1370.

\bibitem[\protect\citename{Zhao \bgroup et al.\egroup }2015]{Zhao15relatedness}
Yu~Zhao, Zhiyuan Liu, and Maosong Sun.
\newblock 2015.
\newblock Representation learning for measuring entity relatedness with rich
  information.
\newblock In {\em Proceedings of the Twenty-Fourth International Joint
  Conference on Artificial Intelligence, {IJCAI} 2015, Buenos Aires, Argentina,
  July 25-31, 2015}, pages 1412--1418.

\bibitem[\protect\citename{Zhou and Zhang}2006]{zhou2006multi}
Zhi{-}Hua Zhou and Min{-}Ling Zhang.
\newblock 2006.
\newblock Multi-instance multi-label learning with application to scene
  classification.
\newblock In {\em Advances in Neural Information Processing Systems 19,
  Proceedings of the Twentieth Annual Conference on Neural Information
  Processing Systems, Vancouver, British Columbia, Canada, December 4-7, 2006},
  pages 1609--1616.

\end{thebibliography}

\end{document}